\documentclass[runningheads]{llncs}
\usepackage[misc]{ifsym}
\usepackage{graphicx} 
\usepackage[normalem]{ulem}
\useunder{\uline}{\ul}{}
\usepackage{graphicx} 
\usepackage[colorlinks=true,allcolors=blue]{hyperref}
\usepackage{amsmath,amssymb,amsfonts,bm}

\usepackage{multirow}
\usepackage{booktabs}
\usepackage{placeins}
\usepackage{subfigure} 
\usepackage{floatrow}
\newfloatcommand{capbtabbox}{table}[][\FBwidth]
\begin{document}
\title{All-In-One Medical Image Restoration via Task-Adaptive Routing}

\author{
Zhiwen Yang\inst{1} \and
Haowei Chen\inst{1} \and 
Ziniu Qian\inst{1} \and 
Yang Yi\inst{1} \and
Hui Zhang\inst{2} \and 
Dan Zhao\inst{3} \and 
Bingzheng Wei\inst{4} \and 
Yan Xu\inst{1}$^{(\textrm{\Letter},\thanks{Corresponding author})}$
} 

\authorrunning{Yang et al.}

\institute{
School of Biological Science and Medical Engineering, State Key Laboratory of Software Development Environment, Key Laboratory of Biomechanics and Mechanobiology of Ministry of Education, Beijing Advanced Innovation Center for Biomedical Engineering, Beihang University, Beijing 100191, China
\\
\email{xuyan04@gmail.com} \and 
Department of Biomedical Engineering, Tsinghua University, Beijing 100084, China \and 
Department of Gynecology Oncology, National Cancer Center/National Clinical Research Center for Cancer/Cancer Hospital, Chinese Academy of Medical Sciences and Peking Union Medical College, Beijing 100021, China \and
ByteDance Inc., Beijing 100098, China
}


%
%
%
\maketitle              
\begin{abstract}

Although single-task medical image restoration (MedIR) has witnessed remarkable success, the limited generalizability of these methods poses a substantial obstacle to wider application. In this paper, we focus on the task of all-in-one medical image restoration, aiming to address multiple distinct MedIR tasks with a single universal model. Nonetheless, due to significant differences between different MedIR tasks, training a universal model often encounters task interference issues, where different tasks with shared parameters may conflict with each other in the gradient update direction. This task interference leads to deviation of the model update direction from the optimal path, thereby affecting the model's performance. To tackle this issue, we propose a task-adaptive routing strategy, allowing conflicting tasks to select different network paths in spatial and channel dimensions, thereby mitigating task interference. Experimental results demonstrate that our proposed \textbf{A}ll-in-one \textbf{M}edical \textbf{I}mage \textbf{R}estoration (\textbf{AMIR}) network achieves state-of-the-art performance in three MedIR tasks: MRI super-resolution, CT denoising, and PET synthesis, both in single-task and all-in-one settings. The code and data will be available at \href{https://github.com/Yaziwel/All-In-One-Medical-Image-Restoration-via-Task-Adaptive-Routing.git}{https://github.com/Yaziwel/AMIR}.

\end{abstract}
\section{Introduction} 
Medical image restoration (MedIR) refers to the process of transforming low-quality (LQ) medical images into high-quality (HQ) images. MedIR has achieved remarkable success in individual tasks such as MRI super-resolution \cite{chen2018mDCSRN,zamir2022restormer}, CT denoising \cite{chen2017ct-cnn,chen2017redcnn,luthra2021eformer,wang2023ctformer}, and PET synthesis \cite{xiang2017auto-contextcnn,chan2018dcnn,luo2022argan,jang2023spach,zhou2022sgsgan,yang2023drmc}. Within the scope of single-task MedIR, there are also studies addressing input degradation variations, such as varying noise levels \cite{chan2018dcnn}, imaging protocols \cite{yang2023drmc}, and centers \cite{yang2023drmc}. The well-defined settings of these MedIR tasks allow researchers to design specific models tailored to the unique characteristics of each individual task.

However, these single-task models designed for particular MedIR tasks often encounter significant performance drops when applied to other MedIR tasks. In complex application scenarios such as multimodal imaging (e.g., PET/CT and PET/MRI), multiple MedIR tasks coexist simultaneously. Single-task models struggle to address the differences between modalities and tasks, and utilizing separate models for each task may result in inefficiencies in both usage and maintenance. Apart from practical limitations, single-task models remain inherently constrained by their task-specific nature, hindering the evolution from specificity to a more generalized intelligence in the field of medical image restoration. Therefore, there is a pressing need to develop a single universal model capable of simultaneously handling multiple MedIR tasks. 

\begin{figure*}[t]
	\centering
	\includegraphics[width=\textwidth]{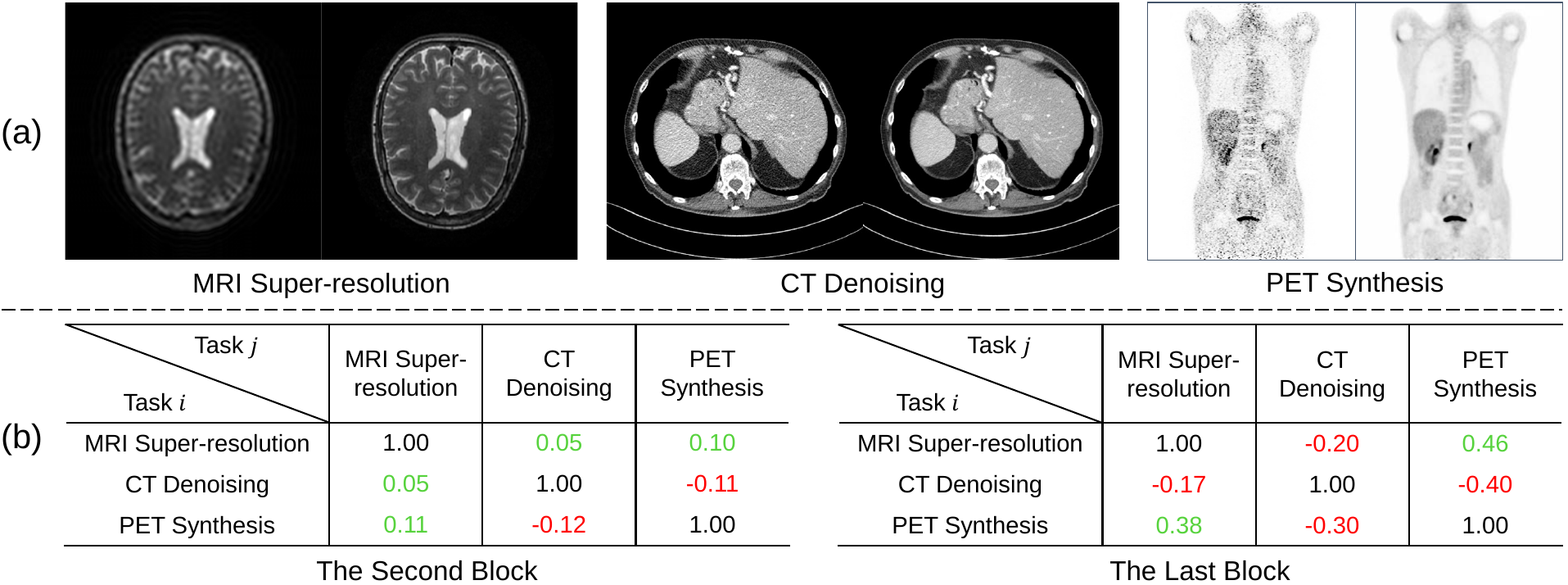}
    \caption{(a) Examples of LQ/HQ pairs for three different MedIR tasks. (b) The interference metric \cite{zhu2022uniperceiver-moe} of task $j$ on task $i$ at the second and last blocks in Restormer \cite{zamir2022restormer}. Red values indicate that task $j$ negatively impacts task $i$, while green values indicate a positive impact.}
	\label{fig:interference}
\end{figure*}

Recently, all-in-one image restoration \cite{li2022airnet,potlapalli2023promptir,park2023discriminative_filter,kong2024mio} (also known as multi-task image restoration) has gained prominence in natural images, attempting to address multiple different restoration tasks using a single universal model. It holds the potential to address multiple MedIR tasks with a single universal model. However, given the substantial disparities between medical and natural image restoration tasks, it is not advisable to directly apply all-in-one natural image restoration methods to MedIR tasks. In natural image restoration, task distinctions primarily arise from varied input image degradations, with their ground truths presumed to follow a uniform data distribution. However, in MedIR, alongside input degradation disparities, the ground truths of different MedIR tasks also showcase significantly varied data distributions due to modality differences (as shown in Fig.~\ref{fig:interference} (a)). These significant differences between MedIR tasks can result in task interference, a common issue in multi-task learning \cite{yu2020gradient_surgery,zhu2022uniperceiver-moe}, where the gradient update directions between tasks are inconsistent or even opposite. As shown in Fig.~\ref{fig:interference} (b), we quantify the interference metric defined in the paper \cite{zhu2022uniperceiver-moe} among different MedIR tasks and observe significant task interference. This task interference issue will lead to an uncertain update direction that deviates from the optimal, resulting in suboptimal model performance. While essential for handling multiple MedIR tasks, the task interference issue is rarely explicitly addressed by current all-in-one methods.

In this paper, we introduce an innovative All-in-one Medical Image Restoration (AMIR) network capable of handling multiple MedIR tasks with a single universal model. The key idea behind AMIR is the incorporation of a task-adaptive routing strategy, which dynamically directs inputs from conflicting tasks to different network paths, explicitly mitigating interference between tasks. Specifically, the proposed task-adaptive routing involves routing instruction learning, spatial routing, and channel routing. Routing instruction learning aims to adaptively learn task-relevant instructions based on input images, while spatial routing and channel routing utilize these learned instructions to guide the routing of network features at spatial and channel levels, respectively, thereby alleviating potential interference. Extensive experiments demonstrate that our proposed AMIR achieves state-of-the-art performance in three MedIR tasks: MRI super-resolution, CT denoising, and PET synthesis, both in single-task and all-in-one settings. Our contribution can be three-fold:

\begin{itemize} 
\item We propose a novel All-in-one Medical Image Restoration (AMIR) network, which allows handling multiple different MedIR tasks with a single unified model. To the best of our knowledge, AMIR could be one of the first methods to handle multiple MedIR tasks in an all-in-one fashion.
\item We propose a novel task-adaptive routing strategy to mitigate interference between different tasks. It is achieved by assigning conflicting tasks to different network paths.
\item Extensive experiments show that our proposed AMIR achieves state-of-the-art performance in both single-task MedIR and all-in-one MedIR tasks. 
\end{itemize}

\begin{figure*}[t]
	\centering
	\includegraphics[width=\textwidth]{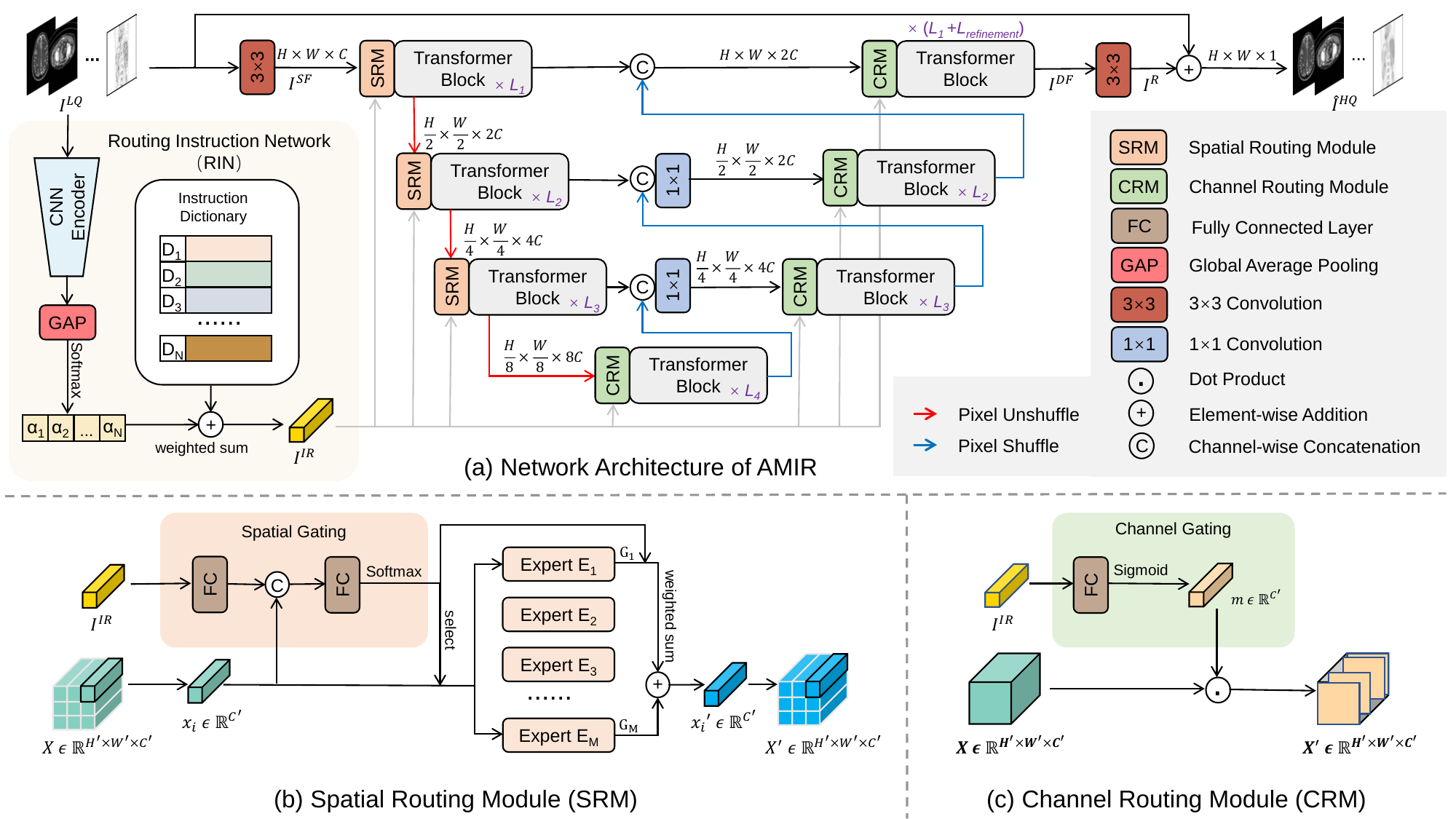}
    \caption{Overview of the proposed all-in-one medical image restoration (AMIR) network.}
	\label{fig:framework}
\end{figure*}

\section{Method}
\subsection{Network Architecture} 
The architecture of the proposed All-In-One Medical Image Restoration (AMIR) network is shown in Fig.~\ref{fig:framework} (a). AMIR adopts Restormer \cite{zamir2022restormer}, a Unet-style network, as the baseline model for medical image restoration. The key difference lies in AMIR's inclusion of a Routing Instruction Network (RIN) alongside Spatial Routing Modules (SRMs; as shown in Fig.~\ref{fig:framework} (b)) and Channel Routing Modules (CRMs; as shown in Fig.~\ref{fig:framework} (c)) for network path routing. Specifically, given an input LQ medical image $I^{LQ} \in \mathbb{R}^{H\times W \times 1}$, AMIR first extracts shallow features $I^{SF} \in \mathbb{R}^{H\times W \times C}$ by applying a 3$\times$3 convolution, where $H$, $W$, and $C$ denote the height, width, and channel, respectively. Subsequently, $I^{SF}$ undergoes a hierarchical encoder-bottleneck-decoder structure to be transformed into deep features $I^{DF}$, with multiple Restormer Transformer blocks \cite{zamir2022restormer} utilized at each level for feature extraction. To mitigate task interference, AMIR employs a task-adaptive routing strategy, incorporating SRMs before each encoder level and CRMs before the bottleneck and each decoder level. SRMs and CRMs adaptively select propagation paths for different task features based on the task-relevant instructions from RIN, thus reducing potential interference. Finally, a 3$\times$3 convolution layer is applied to deep features $I^{DF}$ to generate residual image $I^R \in \mathbb{R}^{H\times W \times 1}$, which is added to the input LQ image to obtain the restored image $\hat{I}^{HQ} = I^{LQ} + I^{R}$. Next, we will describe the task-adaptive routing strategy and its key components in detail.

\subsection{Task-Adaptive Routing} 
To address potential interference caused by multiple tasks sharing the same parameters but with different optimization directions, we propose a task-adaptive routing strategy. This strategy enables different tasks to select distinct paths for customized processing, thereby avoiding interference between tasks. We will introduce the task-adaptive routing strategy from three aspects: routing instruction learning, spatial routing, and channel routing.

\textbf{Routing Instruction Learning.} Instructions are representations relevant to the task, crucial for helping the network better understand the current task and adjust the direction of restoration. Previous all-in-one natural image restoration methods often utilize representations from contrastive learning \cite{he2020contrastive,li2022airnet} as instructions. However, the task of contrastive learning exhibits significant differences from image restoration tasks, making it challenging to balance their relationship and potentially introducing undesirable task interference. To address this, we propose a Routing Instruction Network (RIN; as shown in Fig.~\ref{fig:framework} (a)), which can adaptively generate task-relevant instructions from the input image without the need for additional supervision. The mechanism of the RIN can be formulated as follows:

\begin{equation}
I^{IR}=\sum_{i=1}^N \alpha_{i}D_i, \quad \alpha_i=\operatorname{Softmax}\left(\operatorname{GAP}\left(\operatorname{E}(I)\right)\right),
\end{equation}
where $\operatorname{E(\cdot)}$ is a five-layer CNN encoder, while $\operatorname{GAP(\cdot)}$ denotes the global average pooling layer. $D=[D_1, D_2, ..., D_N]$ constitutes an instruction dictionary, where the $N$ instructions $D_i \in \mathbb{R}^{256}$ in the dictionary are learnable parameters. During this process, RIN dynamically predicts weights $\alpha_i$ from the input image $I$ and applies them to the instruction dictionary $D$ to generate input-conditioned instruction $I^{IR}$. This process does not require any supervision, yet we demonstrate in Fig.~\ref{fig:vis_ablation} (a) that the learned instructions $I^{IR}$ are task-relevant.

\textbf{Spatial Routing.} Task interference arises when different tasks share network parameters but have different update directions. A fundamental solution is to assign conflicting tasks to separate parameters. Mixture-of-Experts (MoE) \cite{shazeer2017moe} provides a potential solution, which learns to dynamically route inputs to different expert networks. However, vanilla MoE selects the experts solely relying on input token representation, neglecting crucial global and task-relevant information. Hence, we propose to enhance the routing strategy of the vanilla MoE by incorporating the learned global task-relevant instruction $I^{IR}$. Given a local spatial token $x_i \in \mathbb{R}^{C'}$ of the input feature $X \in \mathbb{R}^{H'\times W'\times C'}$, along with the global instruction $I^{IR}$ and an expert bank $\operatorname{E}=[\operatorname{E}_1, \operatorname{E}_2, ..., \operatorname{E}_M]$ comprising $M$ expert networks, our spatial routing module (SRM; as shown in Fig.~\ref{fig:framework} (b)) can be formulated as:
\begin{equation}
\operatorname{G}(x_i, I^{IR})=\operatorname{Top-K}(\operatorname{Softmax}\left(\operatorname{FC}\left([x_i, \operatorname{FC}(I^{IR})]\right)\right)),
\end{equation} 

\begin{equation}
x_{i}'=\sum_{e=1}^M \operatorname{G}(x_i, I^{IR})_e\operatorname{E}_e\left(x_i\right),
\end{equation} 
where $\operatorname{FC}(\cdot)$ indicates the fully connected layer. $\operatorname{Top-K}(\cdot)$ operator sets all values to be zero except the largest $K$ values. $\operatorname{G}(\cdot)$ denotes the routing function that produces a sparse weight for different experts. $\operatorname{E}_e$ refers to the $e$-th expert in the expert bank, with each being a multi-layer perception (MLP). $\operatorname{G}(x_i, I^{IR})_e$ determines how much the $e$-th expert contributes to the output. In this process, SRM routes each spatial token from the input feature $X$ to the corresponding top-$K$ selected experts for separate processing with the guidance of the global instruction $I^{IR}$, and then combines each expert's knowledge through weighted summation. Notably, very dissimilar tokens will be routed to distinct experts, thereby avoiding potential interference. We integrate SRM before each encoder level in the UNet to mitigate task interference during the encoding process.

\textbf{Channel Routing.} Although SRM can effectively mitigate task interference and obtain strong interpretability, it suffers from two drawbacks. Firstly, routing different spatial tokens to different experts disrupts the feature spatial continuity. Secondly, SRM introduces multiple expert networks, leading to a linear increase in parameters with the growth of expert networks. To address these issues, we propose a more efficient Channel Routing Module (CRM; as shown in Fig.~\ref{fig:framework} (c)). With the guidance of task instruction $I^{IR}$, CRM dynamically routes data flow from different tasks to different channels without introducing excessive additional parameters, while also preserving feature spatial continuity. Given an input feature $X \in \mathbb{R}^{H'\times W'\times C'}$, the mechanism of CRM can be described as follows:
\begin{equation}
X'=(X\odot m), \quad m = \operatorname{Sigmoid}(\operatorname{FC}(I^{IR})),
\end{equation} 
where $m\in \mathbb{R}^{C'}$ is a channel-wise (soft-)binary mask estimated from the instruction $I^{IR}$. In this process, CRM conducts channel-wise task routing through feature masking depending on the task instruction. We incorporate the CRM before the bottleneck and each decoder level of the UNet to address task interference in the decoding process. 

\subsection{Loss Function} 
The overall loss can be summarized as follows:
\begin{equation}
L = L_1 + \gamma L_{Balance},
\end{equation} 
where $L_1$ depicts the difference between the restored $\hat{I}^{HQ}$ and the high-quality image $I^{HQ}$. $L_{Balance}$ \cite{shazeer2017moe} is a regularization term in MoE that prevents static routing where the same few experts are always selected. $\gamma$ is a weighting parameter.

\section{Experiments and Results} 
\subsection{Dataset} 
Our all-in-one medical image restoration experiment encompasses three tasks: MRI super-resolution, CT denoising, and PET Synthesis. In the subsequent section, we introduce the corresponding datasets for each task.

\textbf{MRI Super-Resolution.} We use the publicly available IXI MRI dataset \cite{ixi_dataset}, which comprises 578 HQ T2 weighted MRI images. Each 3D MRI volume is sized at 256$\times$256$\times n$, from which we extract the central 100 2D slices sized 256$\times$256 to exclude side slices. The LQ image is obtained by cropping the $k$-space with a downsampling factor of 4$\times$ (retaining the central 6.25$\%$ data points). The dataset is divided into 405 for training, 59 for validation, and 114 for testing.

\textbf{CT Denoising.} We employ the dataset from the 2016 NIH AAPM-Mayo Clinic Low-Dose CT Grand Challenge \cite{mccollough2017AAPM}, which comprises paired standard-dose HQ CT images and quarter-dose LQ CT images, each with an image size of 512$\times$512. These images originate from 10 patients, with 8 allocated for training, 1 for validation, and 1 for testing purposes.

\textbf{PET Synthesis.} We acquire 159 HQ PET images using the PolarStar m660 PET/CT system in list mode, with an injection dose of 293MBq $^{18}$F-FDG. LQ PET images are generated through random list mode decimation with a dose reduction factor of 12. Both HQ and LQ PET images are reconstructed using the standard OSEM method \cite{hudson1994osem}. Each PET image has 3D shapes of 192$\times$192$\times$400, with a voxel size of 3.15mm$\times$3.15mm$\times$1.87mm, and is divided into 192 2D slices sized 192$\times$400. Slices containing only air are excluded. Patient data are partitioned into 120 for training, 10 for validation, and 29 for testing. 

\subsection{Implementation} 
In our AMIR architecture, as shown in Fig.~\ref{fig:framework}, the number of Transformer blocks is configured as follows: $L_1=5$, $L_2=L_3=7$, $L_4=9$, and $L_{refinement}=4$. The input channel number is specified as $C=42$, and the length of the instruction dictionary is set to $N=16$. Within the SRM component, we designate the expert number as $M=4$ and the selected number of experts as $K=2$. During training, we utilize patches sized at $128\times128$ with a batch size of 8. Our model undergoes training via the Adam optimizer for $2\times 10^{5}$ iterations, starting with an initial learning rate of $2\times10^{-4}$, gradually decreasing to $1\times10^{-6}$ using cosine annealing. All experiments are conducted in PyTorch, utilizing NVIDIA A100 with 40GB memory. 

\begin{table}[t] 
\centering
\caption{Single task medical image restoration results. The best results are \textbf{bolded}, and the second-best results are {\ul underlined}.} 
\resizebox{\textwidth}{!}{
\begin{tabular}{c|ccc|c|ccc|c|ccc}
\hline
\multirow{2}{*}{Method} & \multicolumn{3}{c|}{MRI Super-Resolution}             & \multirow{2}{*}{Method} & \multicolumn{3}{c|}{CT Denoising}                    & \multirow{2}{*}{Method} & \multicolumn{3}{c}{PET Synthesis}                    \\ \cline{2-4} \cline{6-8} \cline{10-12} 
                        & PSNR↑            & SSIM↑           & RMSE↓            &                         & PSNR↑            & SSIM↑           & RMSE↓           &                         & PSNR↑            & SSIM↑           & RMSE↓           \\ \hline
SRCNN \cite{dong2015srcnn}                   & 28.8067          & 0.8919          & 41.3488          & CNN \cite{chen2017ct-cnn}                    & 32.7600          & 0.9075          & 9.3928          & Xiang's \cite{xiang2017auto-contextcnn}                & 35.9268          & 0.9167          & 0.0980          \\
VDSR \cite{kim2016vdsr}                    & 30.0446          & 0.9140          & 36.0508          & REDCNN \cite{chen2017redcnn}                 & 33.1889          & 0.9113          & 8.9427          & DCNN \cite{chan2018dcnn}                   & 36.2710          & 0.9243          & 0.0954          \\
SwinIR \cite{liang2021swinir}                 & 31.5549          & 0.9334          & 30.5788         & Eformer \cite{luthra2021eformer}                 & {\ul 33.3496}    & {\ul 0.9175}    & {\ul 8.8030}    & ARGAN \cite{luo2022argan} & 36.7272          & 0.9406          & 0.0902          \\
Restormer \cite{zamir2022restormer}               & {\ul 31.8474}    & {\ul 0.9378}    & {\ul 29.7005}    & CTformer \cite{wang2023ctformer}               & 33.2506          & 0.9134          & 8.8974          & SpachTransformer \cite{jang2023spach}        & {\ul 37.1371}    & {\ul 0.9456}    & {\ul 0.0871}    \\
AMIR                    & \textbf{31.9923} & \textbf{0.9393} & \textbf{29.2095} & AMIR                    & \textbf{33.6738} & \textbf{0.9183} & \textbf{8.4773} & AMIR                    & \textbf{37.2121} & \textbf{0.9473} & \textbf{0.0863} \\ \hline
\end{tabular}
}
\label{tab:single_task}
\end{table}

\begin{figure*}[t]
	\centering
	\includegraphics[width=\textwidth]{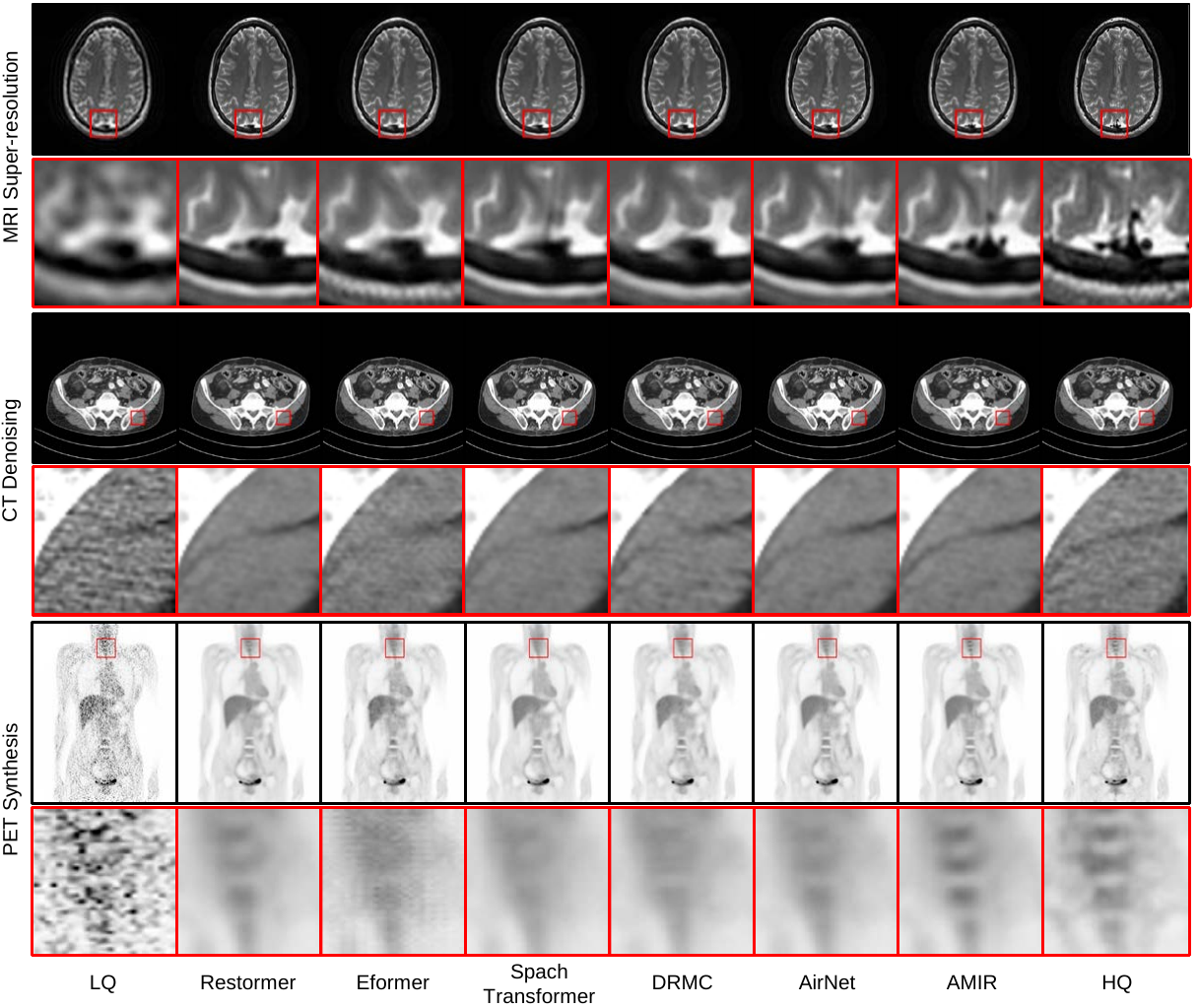}
    \caption{Visual comparison of different methods on all-in-one medical image restoration.}
	\label{fig:vis_comaparison}
\end{figure*}

\begin{table}[t] 
\centering
\caption{All-in-one medical image restoration results.} 
\resizebox{\textwidth}{!}{
\begin{tabular}{c|ccc|ccc|ccc|ccc}
\hline
\multirow{2}{*}{Method} & \multicolumn{3}{c|}{MRI Super-Resolution}             & \multicolumn{3}{c|}{CT Denoising}                    & \multicolumn{3}{c|}{PET Synthesis}                   & \multicolumn{3}{c}{Average}                           \\ \cline{2-13} 
                        & PSNR↑            & SSIM↑           & RMSE↓            & PSNR↑            & SSIM↑           & RMSE↓           & PSNR↑            & SSIM↑           & RMSE↓           & PSNR↑            & SSIM↑           & RMSE↓            \\ \hline
Restormer \cite{zamir2022restormer}               & {\ul 31.7177}    & {\ul 0.9362}    & {\ul 30.0549}    & 33.6142          & {\ul 0.9177}          & 8.5329          & {\ul 37.1368}    & {\ul 0.9473}    & {\ul 0.0872}    & {\ul 34.1562}    & {\ul 0.9337}    & {\ul 12.8917}    \\
Eformer \cite{luthra2021eformer}                 & 29.1922          & 0.8728          & 39.0983          & 32.4438          & 0.9078          & 9.7565          & 35.1096          & 0.9091          & 0.1085          & 32.2485          & 0.8966          & 16.3211          \\
Spach Transformer \cite{jang2023spach}       & 31.1799          & 0.9290          & 31.8342          & 33.4740          & 0.9155          & 8.6677          & 37.0547          & 0.9445          & 0.0874          & 33.9029          & 0.9297          & 13.5298          \\
DRMC \cite{yang2023drmc}                   & 29.5466          & 0.9032          & 38.1691          & 33.2770          & 0.9153          & 8.8674          & 36.1909          & 0.9376          & 0.0960          & 33.0048          & 0.9187          & 15.7108          \\
AirNet \cite{li2022airnet}                  & 31.3921          & 0.9316          & 31.1141          & {\ul 33.6222}    &  0.9176    & {\ul 8.5226}    & \textbf{37.1721} & 0.9451          & \textbf{0.0864} & 34.0621          & 0.9314          & 13.2410          \\
AMIR                    & \textbf{32.0262} & \textbf{0.9396} & \textbf{29.0988} & \textbf{33.7011} & \textbf{0.9182} & \textbf{8.4520} & 37.1193          & \textbf{0.9475} & 0.0876          & \textbf{34.2822} & \textbf{0.9351} & \textbf{12.5461} \\ \hline
\end{tabular}
}
\label{tab:all_in_one_comparison}
\end{table}  

\subsection{Comparative Experiment} 
To validate our proposed AMIR, we conduct evaluations on three tasks: MRI super-resolution, CT denoising, and PET synthesis. Baselines for MRI super-resolution include SRCNN \cite{dong2015srcnn}, VDSR \cite{kim2016vdsr}, SwinIR \cite{liang2021swinir}, and Restormer \cite{zamir2022restormer}; for CT denoising, CNN \cite{chen2017ct-cnn}, REDCNN \cite{chen2017redcnn}, Eformer \cite{luthra2021eformer}, and CTformer \cite{wang2023ctformer}; and for PET synthesis, Xiang's method \cite{xiang2017auto-contextcnn}, DCNN \cite{chan2018dcnn}, ARGAN \cite{luo2022argan}, and Spach Transformer \cite{jang2023spach}. We conduct experiments under two settings: single-task and all-in-one. In the single-task setting, we train different models for each MedIR task and compare them with their respective baselines. In the all-in-one setting, we train a universal model to address all tasks simultaneously. We retrain the best-performing baseline models from each task into the all-in-one setting for comparison. Additionally, we utilize two universal models for comparison:  DRMC \cite{yang2023drmc}, initially developed for multi-center PET synthesis, and AirNet \cite{li2022airnet}, originally designed for all-in-one natural image restoration. PSNR, SSIM, and RMSE scores are calculated to assess the restoration performance.

\textbf{Single-Task Medical Image Restoration.} The results of single-task MedIR are showcased in Table.~\ref{tab:single_task}, revealing that our proposed AMIR surpasses the best-performing baseline models — Restormer \cite{zamir2022restormer}, Eformer \cite{luthra2021eformer}, and Spach Transformer \cite{jang2023spach} — in MRI super-resolution, CT denoising, and PET synthesis tasks, respectively. Despite not being specifically designed for single-task MedIR, AMIR's outstanding performance suggests that the proposed routing strategy is also helpful for handling the sample differences within a single task.

\textbf{All-In-One Medical Image Restoration.} The results of all-in-one MedIR are presented in Table.~\ref{tab:all_in_one_comparison}, where AMIR demonstrates state-of-the-art performance when averaged across the three tasks. Specifically, in MRI super-resolution and CT denoising, AMIR outperforms all comparative methods in terms of PSNR, SSIM, and RMSE metrics. In PET synthesis, although AMIR's PSNR and RMSE metrics are lower than AirNet \cite{li2022airnet}, it achieves the best result in SSIM. Visual comparisons in Fig.~\ref{fig:vis_comaparison} demonstrate that AMIR better restores image structure and details across all three tasks. The superiority of AMIR over other methods lies in its ability to mitigate task interference more effectively, thereby preserving the specificity of each task. 


\subsection{Ablation Study} 

We conduct ablation studies on different training task combinations to analyze their impact on AMIR outcomes. Also, we study the task-adaptive routing strategy through ablation studies to understand the specific roles of its components.

\textbf{Ablation Study on Task Combinations.} We train AMIR on various task combinations and list the results in Table.~\ref{tab:ablation_task_combination}. It is surprising that, despite the increase in the number of tasks, AMIR maintains its performance without significant degradation. This can be attributed to the proposed task-adaptive routing strategy, which enables multiple tasks to share a universal model with minimal interference. Furthermore, Table.~\ref{tab:ablation_task_combination} reveals that certain task combinations yield better results than single-task models. This is because universal models benefit from more training data and task synergy compared to single-task models.

\begin{table}[t] 
\centering
\caption{Ablation study on the combinations of training tasks. "$\checkmark$" denotes AMIR training with the task and "$\textendash$" indicates unavailable results.} 
\resizebox{\textwidth}{!}{
\begin{tabular}{ccc|ccccccccc}
\hline
\multicolumn{3}{c|}{Training Task}                                                                     & \multicolumn{9}{c}{Testing Task}                                                                                                                                                                              \\ \hline
\multirow{2}{*}{MRI Super-Resolution} & \multirow{2}{*}{CT Denoising} & \multirow{2}{*}{PET Synthesis} & \multicolumn{3}{c|}{MRI Super-Resolution}                                  & \multicolumn{3}{c|}{CT Denoising}                                         & \multicolumn{3}{c}{PET Synthesis}                    \\ \cline{4-12} 
                                      &                               &                                & PSNR↑            & SSIM↑           & \multicolumn{1}{c|}{RMSE↓}            & PSNR↑            & SSIM↑           & \multicolumn{1}{c|}{RMSE↓}           & PSNR↑            & SSIM↑           & RMSE↓           \\ \hline
$\checkmark$                          &                               &                                & 31.9923          & 0.9393          & \multicolumn{1}{c|}{29.2095}          & $\textendash$    & $\textendash$   & \multicolumn{1}{c|}{$\textendash$}   & $\textendash$    & $\textendash$   & $\textendash$   \\
                                      & $\checkmark$                  &                                & $\textendash$    & $\textendash$   & \multicolumn{1}{c|}{$\textendash$}    & 33.6738          & {\ul 0.9183}    & \multicolumn{1}{c|}{8.4773}          & $\textendash$    & $\textendash$   & $\textendash$   \\
                                      &                               & $\checkmark$                   & $\textendash$    & $\textendash$   & \multicolumn{1}{c|}{$\textendash$}    & $\textendash$    & $\textendash$   & \multicolumn{1}{c|}{$\textendash$}   & \textbf{37.2121} & {\ul 0.9473}    & \textbf{0.0863} \\
$\checkmark$                          & $\checkmark$                  &                                & \textbf{32.0683} & \textbf{0.9404} & \multicolumn{1}{c|}{\textbf{29.0296}} & 33.6934          & {\ul 0.9183}    & \multicolumn{1}{c|}{8.4592}          & $\textendash$    & $\textendash$   & $\textendash$   \\
$\checkmark$                          &                               & $\checkmark$                   & {\ul 32.0404}    & {\ul 0.9399}    & \multicolumn{1}{c|}{{\ul 29.0864}}    & $\textendash$    & $\textendash$   & \multicolumn{1}{c|}{$\textendash$}   & 37.1054          & {\ul 0.9473}    & {\ul 0.0875}    \\
                                      & $\checkmark$                  & $\checkmark$                   & $\textendash$    & $\textendash$   & \multicolumn{1}{c|}{$\textendash$}    & \textbf{33.7537} & \textbf{0.9189} & \multicolumn{1}{c|}{\textbf{8.4027}} & 37.0738          & {\ul 0.9473}    & 0.0880          \\
$\checkmark$                          & $\checkmark$                  & $\checkmark$                   & 32.0262          & 0.9396          & \multicolumn{1}{c|}{29.0988}          & {\ul 33.7011}    & 0.9182          & \multicolumn{1}{c|}{{\ul 8.4520}}    & {\ul 37.1193}    & \textbf{0.9475} & 0.0876          \\ \hline
\end{tabular}
}
\label{tab:ablation_task_combination}
\end{table}  

\begin{figure*}[t]
	\centering
	\includegraphics[width=\textwidth]{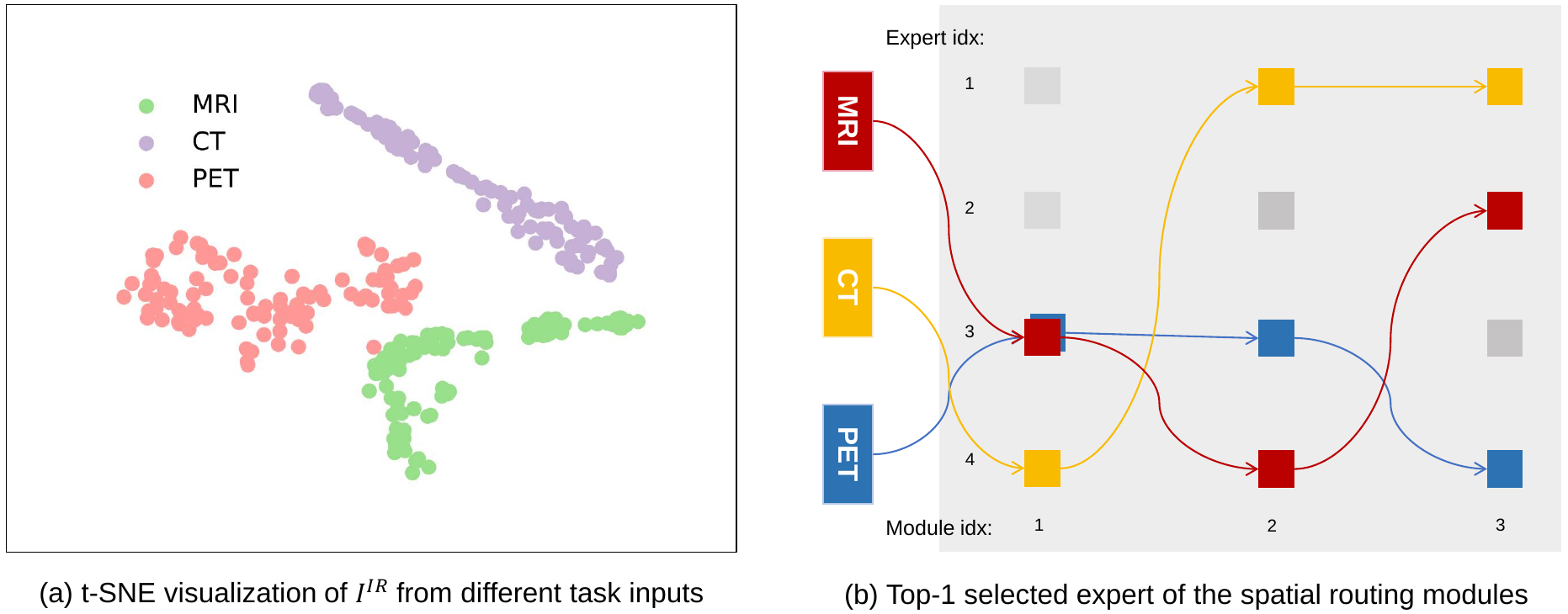}
    \caption{(a) t-SNE visualization of \(I^{IR}\) from different tasks, indicating a clear clustering of different task inputs. (b) Top-1 selected expert in each spatial routing module (SRM). In our AMIR network setting, there are 3 SRMs, each incorporating a mixture of experts (MOE) with 4 experts. Within these SRMs, the top-1 selected expert is identified across the 3 SRMs for each task. Remarkably, the top experts selected across SRMs form distinct paths, with variations observed across different tasks.}
	\label{fig:vis_ablation}
\end{figure*}

\textbf{Ablation Study on the Task-Adaptive Routing Strategy.} 
We conduct ablations on instruction learning and routing modules to analyze the components of the task-adaptive routing strategy. For instruction learning, we remove the dictionary $D$ and adopt contrastive learning \cite{he2020contrastive}, similar to AirNet \cite{li2022airnet}, for guidance. Table.~\ref{tab:ablation_component_analysis} demonstrates the superior effectiveness of adaptive instruction learning. The instruction representation $I^{IR}$ is visualized in Fig.~\ref{fig:vis_ablation} (a) using t-SNE, revealing discriminations between tasks, thus demonstrating its relevance to tasks. Regarding the routing modules SRM and CRM, we assess their effectiveness by removing them individually. Table.~\ref{tab:ablation_task_combination} indicates that both of them effectively enhance the model's performance. Additionally, it is shown in Fig.~\ref{fig:vis_ablation} (b) that different tasks select distinct expert networks within SRM, confirming its interpretability. Thanks to the well-designed task-adaptive routing strategy, our proposed AMIR achieves better results than the baseline model Restormer \cite{zamir2022restormer} even with fewer parameters (although introducing additional parameters, AMIR utilizes fewer channels).

\begin{table}[t] 
\centering
\caption{Ablation study on the task-adaptive routing strategy.} 
\resizebox{\textwidth}{!}{
\begin{tabular}{cc|c|ccc|ccc|ccc}
\hline
\multicolumn{2}{c|}{\multirow{2}{*}{Method}}                                               & \multirow{2}{*}{Params} & \multicolumn{3}{c|}{MRI Super-Resolution}             & \multicolumn{3}{c|}{CT Denoising}                    & \multicolumn{3}{c}{PET Synthesis}                    \\ \cline{4-12} 
\multicolumn{2}{c|}{}                                                                      &                         & PSNR↑            & SSIM↑           & RMSE↓            & PSNR↑            & SSIM↑           & RMSE↓           & PSNR↑            & SSIM↑           & RMSE↓           \\ \hline
\multicolumn{2}{c|}{Restormer \cite{zamir2022restormer}   (Baseline)}                                                & 26.12 M                 & 31.7177          & 0.9362          & 30.0549          & 33.6142          & 0.9177          & 8.5329          & 37.1368          & 0.9473          & {\ul 0.0872}    \\ \hline
\multicolumn{1}{c|}{\multirow{2}{*}{Instruction}}    & w/o $D$                               & 23.53 M                 & 31.9088          & 0.9382          & 29.4575          & 33.6800          & {\ul 0.9181}    & {\ul 8.4712}    & 37.1233          & 0.9474          & 0.0876          \\
\multicolumn{1}{c|}{}                                & w/o $D$ and w/ Contrastive   Learning \cite{he2020contrastive} & 23.53 M                 & 31.9545          & 0.9388          & 29.3359          & 33.5987          & 0.9161          & 8.5495          & 37.1388          & 0.9465          & {\ul 0.0872}    \\ \hline
\multicolumn{1}{c|}{\multirow{2}{*}{Routing Module}} & w/o SRM                             & 22.74 M                 & 31.8678          & 0.9379          & 29.5851          & 33.6556          & 0.9171          & 8.4942          & \textbf{37.1639} & 0.9470          & \textbf{0.0871} \\
\multicolumn{1}{c|}{}                                & w/o CRM                             & 23.37 M                 & {\ul 31.9802}    & {\ul 0.9391}    & {\ul 29.2251}    & {\ul 33.6814}    & {\ul 0.9181}    & {\ul 8.4712}    & {\ul 37.1443}    & \textbf{0.9476} & 0.0874          \\ \hline
\multicolumn{2}{c|}{AMIR}                                                                  & 23.54 M                 & \textbf{32.0262} & \textbf{0.9396} & \textbf{29.0988} & \textbf{33.7011} & \textbf{0.9182} & \textbf{8.4520} & 37.1193          & {\ul 0.9475}    & 0.0876          \\ \hline
\end{tabular}
}
\label{tab:ablation_component_analysis}
\end{table}  

\section{Conclusion}
In this paper, we propose an all-in-one medical image restoration (AMIR) network capable of handling multiple MedIR tasks with a single universal model. To mitigate task interference, we introduce a task-adaptive routing strategy that dynamically routes different tasks to distinct network paths. Experiments demonstrate that the proposed AMIR achieves state-of-the-art performance in both single-task MedIR and all-in-one MedIR tasks. In the future, we will explore the effectiveness of the proposed AMIR as more MedIR tasks are involved.

\section{Acknowledgment}
This work is supported by the National Natural Science Foundation in China under Grant 62371016, U23B2063, 62022010, and 62176267, the Bejing Natural Science Foundation Haidian District Joint Fund in China under Grant L222032, the Beijing hope run special fund of cancer foundation of China under Grant LC2018L02, the Fundamental Research Funds for the Central University of China from the State Key Laboratory of Software Development Environment in Beihang University in China, the 111 Proiect in China under Grant B13003, the SinoUnion Healthcare Inc. under the eHealth program, the high performance computing (HPC) resources at Beihang University.
%
%
%
%
\bibliographystyle{splncs04}
\bibliography{ref}
\end{document}